\documentclass[letterpaper]{article} 
\usepackage{aaai2026}  
\usepackage{times}  
\usepackage{helvet}  
\usepackage{courier}  
\usepackage[hyphens]{url}  
\usepackage{graphicx} 
\usepackage{natbib}  
\usepackage{caption} 
\usepackage{algorithm}
\usepackage{algorithmic}
\usepackage{epsfig}
\usepackage{graphicx}
\usepackage{amsmath}
\usepackage{amssymb}
\usepackage{bm}
\usepackage{multirow}
\usepackage{color}
\usepackage{booktabs} 
\usepackage{pifont}

\newcommand{\cmark}{\ding{51}}%
\newcommand{\figref}[1]{Fig.~\ref{#1}}%
\newcommand{\tabref}[1]{Table~\ref{#1}}%

\renewcommand{\eqref}[1]{Equation~(\ref{#1})}

\usepackage{newfloat}
\usepackage{listings}
\DeclareCaptionStyle{ruled}{labelfont=normalfont,labelsep=colon,strut=off} 
\floatstyle{ruled}
\newfloat{listing}{tb}{lst}{}
\floatname{listing}{Listing}
\title{CareCom: Generative Image Composition with Calibrated Reference Features}
\author{
    Jiaxuan Chen\textsuperscript{\rm 1}, Bo Zhang\textsuperscript{\rm 1}, Qingdong He\textsuperscript{\rm 2}, Jinlong Peng\textsuperscript{\rm 2}, Li Niu\textsuperscript{\rm 1,3}\thanks{Corresponding Author.}
}
\affiliations{
    \textsuperscript{\rm 1}MoE Key Lab of Artificial Intelligence, Shanghai Jiao Tong University\\
    \textsuperscript{\rm 2}Youtu Lab, Tencent   \quad \textsuperscript{\rm 3}miguo.ai\\
    \{chenjiaxuan, bo-zhang, ustcnewly\}@sjtu.edu.cn,
    \{yingcaihe, jeromepeng\}@tencent.com
}

\usepackage{bibentry}

\begin{document}

\maketitle

\begin{abstract}

Image composition aims to seamlessly insert foreground object into background. Despite the huge progress in  generative image composition, the existing methods are still struggling with simultaneous detail preservation and foreground
pose/view adjustment. To address this issue, we extend the existing generative composition model to multi-reference version, which allows using arbitrary number of foreground reference images. 
Furthermore, we propose to calibrate the global and local features of foreground reference images to make them compatible with the background information. The calibrated reference features can supplement the original reference features with useful global and local information of proper pose/view. 
Extensive experiments on MVImgNet and MureCom demonstrate that the generative model can greatly benefit from the calibrated reference features.  

\end{abstract}


\section{Introduction}
\label{sec:intro}
Image composition is an important image editing operation, aiming to seamlessly
insert a given foreground object into a background image. Previous 
methods~\cite{tsai2017deep,tsai2017deepblending,hong2022shadow} attempted to
address different issues in image composition with different sub-tasks (\emph{e.g.}, image blending, image harmonization, and shadow generation). 
Recently, foundation diffusion models~\cite{stablediffusion,esser2024scaling,Flux} 
have demonstrated powerful image generation ability, and some works have utilized such ability to re-generate the foreground in the background image with all the issues solved simultaneously. 

\begin{figure}[t]
\centering
\includegraphics[width=\linewidth]{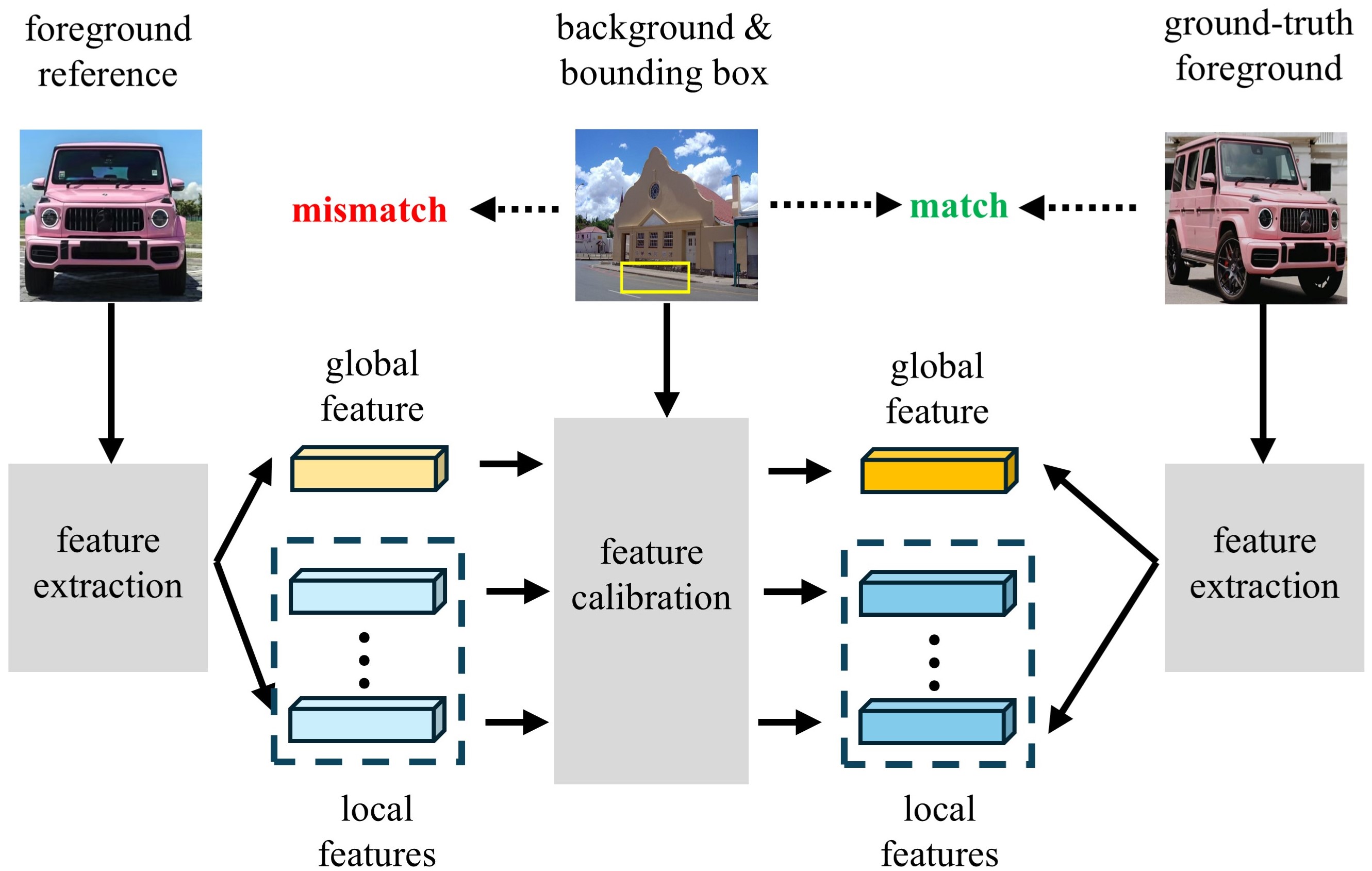} 
\caption{Illustration of feature calibration process. Based on the background and foreground bounding box, we calibrate the features of foreground reference images to match those of ground-truth foreground.}
\label{tease}
\end{figure}

These generative composition methods can be roughly classified into 
training-free methods and training-based methods. 
The training-free methods~\cite{lu2023tf,primecomposer,xu2025context} leverages the prior knowledge in foundation model without the need of training or finetuning. However, they cannot adjust the pose/view of foreground and the generated images are of low quality. 
In contrast, training-based methods~\cite{chen2024anydoor,yang2023paint,
lu2023dreamcom,kulal2023affordance,zhang2023controlcom,winter2024objectmate,winter2024objectdrop,canet2025thinking,yuan2023customnet,song2024imprint,chen2025freecompose} are more popular and more powerful. They require large-scale training set to train the model or a few images containing the target object to finetune the model. The usual approach \cite{Song_2023_CVPR,yang2023paint} is extracting foreground information and injecting into denoising UNet of stable diffusion~\cite{stablediffusion}. 
Some subsequent works attempted to better preserve the foreground details by using different strategies, like high-frequency information \cite{chen2024anydoor} or local enhancement module \cite{zhang2023controlcom}. 

Despite the remarkable progress achieved for generative composition, based on our experimental observation, there is no method adept at both detail preservation and pose/view adjustment of foreground. 
One reason is that the existing methods only use one reference image for the foreground object, which raises the difficulty of generating the details of dramatically different pose/view. 
To explore the advantage of using multiple reference images, we extend the existing generative composition model \cite{Song_2023_CVPR} to support arbitrary number of reference images, by simply concatenating their reference features. 
Based on the multi-reference generative composition model, we observe that using multiple reference images for the foreground object can greatly improve the results, because the model can attend to the reference image with the pose/view relatively matching the background, which can greatly alleviate the task difficulty.
Ideally, if the model is provided with abundant reference images covering all views and poses, the model can attend to the one perfectly matching the background.
However, it is very costly and even impossible to collect a large number of reference images covering the full range of poses/views.

Another thought is to hallucinate the reference image matching the background based on the given reference images. Instead of hallucinating the compatible reference image, we opt for hallucinating the compatible reference feature, \emph{i.e.}, the feature of compatible reference image,
which is more efficient by omitting the transformation process between feature space and image space. Specifically, we design a feature calibration module. This module takes in one reference feature and produces its calibrated reference feature, which is calibrated towards the compatible reference feature based on the background information.   
We refer to the calibrated reference features as augmented reference features, which are appended to the original reference features and jointly sent to denoising UNet. Among the original reference features, for those relatively compatible with the background, their calibrated versions are expected to be more compatible with the background and the model is expected to attend to these calibrated reference features. 

When calibrating the reference features, we consider both global reference features and local reference features, by using global calibration module and local calibration module respectively. The global calibration module relies on background information and current denoised foreground to calibrate the global reference feature, which should be close to the global feature of ground-truth foreground. Similarly, the local calibration module calibrates the local reference feature, which should be close to the corresponding local feature of ground-truth foreground. We name our image \textbf{com}position method with \textbf{ca}librated \textbf{re}ference features as CareCom.

We conduct experiments on MVImgNet~\cite{yu2023mvimgnet} and MureCom~\cite{lu2023dreamcom} 
datasets. They contain multiple reference images for each foreground object, which is suitable for our task. The results demonstrate that our CareCom excels in detail preservation and pose/view adjustment of foreground at the same time. Our contributions can be summarized as follows:
1) We propose the first multi-reference generative composition model supporting an arbitrary number of foreground reference images.
2) We propose to calibrate the foreground reference features to match the background. Technically, we design a global (\emph{resp.}, local) calibration 
module to calibrate the global (\emph{resp.}, local) reference features.
3) Comprehensive experiments on two datasets show that our 
    method outperforms other baselines in terms of faithfulness and realism.
\section{Related Work}
\label{sec:related_works}
\subsection{Generative Image Composition}

In recent years, generative composition has emerged for object insertion with one unified model, thanks to the unprecedented potential of foundation generative model like stable diffusion~\cite{stablediffusion} and Diffusion transformer~\cite{esser2024scaling,Flux}. 
These methods can be categorized into training-based~\cite{yang2023paint,lu2023dreamcom,kulal2023affordance,
winter2024objectdrop,song2024imprint,chen2024anydoor,canet2025thinking,Tarre2025Multitwine,huang2025dreamfuse,yu2025omnipaint}  and training-free approaches~\cite{lu2023tf,primecomposer,pham2024tale,li2024tuning,xu2025context}. 
Our method belongs to the training-based group, which has much stronger ability in adjusting the pose and view of foreground object. 
Among the existing training-based methods, some methods attempt to promote the foreground details \cite{chen2024anydoor,zhang2023controlcom,Song_2023_CVPR} or impose additional controls \cite{zhang2023controlcom}. Some methods \cite{winter2024objectdrop,canet2025thinking} place emphasis on shadow and reflection generation. Some more recent works \cite{song2025insert,wang2025unicombine} explore in-context learning or multi-condition control based on DiT architecture.
However, the above methods only support one reference image, which limits the performance upper bound when multiple reference images are available.

\subsection{Subject-driven Image Generation and Editing}

Subject-driven image generation refers to a variety of tasks of generating or editing images in terms of specific object.

With the emergence of diffusion models, many works~\cite{kumari2023multi,ruiz2024hyperdreambooth,ruiz2023dreambooth,textualinversion} 
have explored text-based image customization. They propose to learn new concepts 
by associating specific object with special text token. 
However, these methods are only applicable to text-generated background instead of a given background image.
Some other works~\cite{wei2023elite,gal2023encoder,tao2025instantcharacter} 
suggest using specific encoders to extract visual information and integrating 
it into customized images. However, they cannot control the placement 
of the foreground. Some approaches~\cite{photoswap,mokady2023null,choi2023custom,yang2024imagebrush,li2025iccustom,
zhang2025ICEdit} 
use text or image as guidance for image editing. 
Despite the diversity of subject-driven tasks, our method primarily focuses on generative composition, re-generating compatible foreground object at the designated location in the background image. 
\section{Our Method}
\label{sec:method}

In Section~\ref{sec:multi_ref_com}, we will introduce our multi-reference generative composition framework. In Section~\ref{sec:global_cal} and \ref{sec:local_cal}, we will elaborate on our global and local reference feature calibration modules respectively. In Section~\ref{sec:training_strategy}, we will introduce the training strategy. 

\begin{figure*}[t]
\centering
\includegraphics[width=\linewidth]{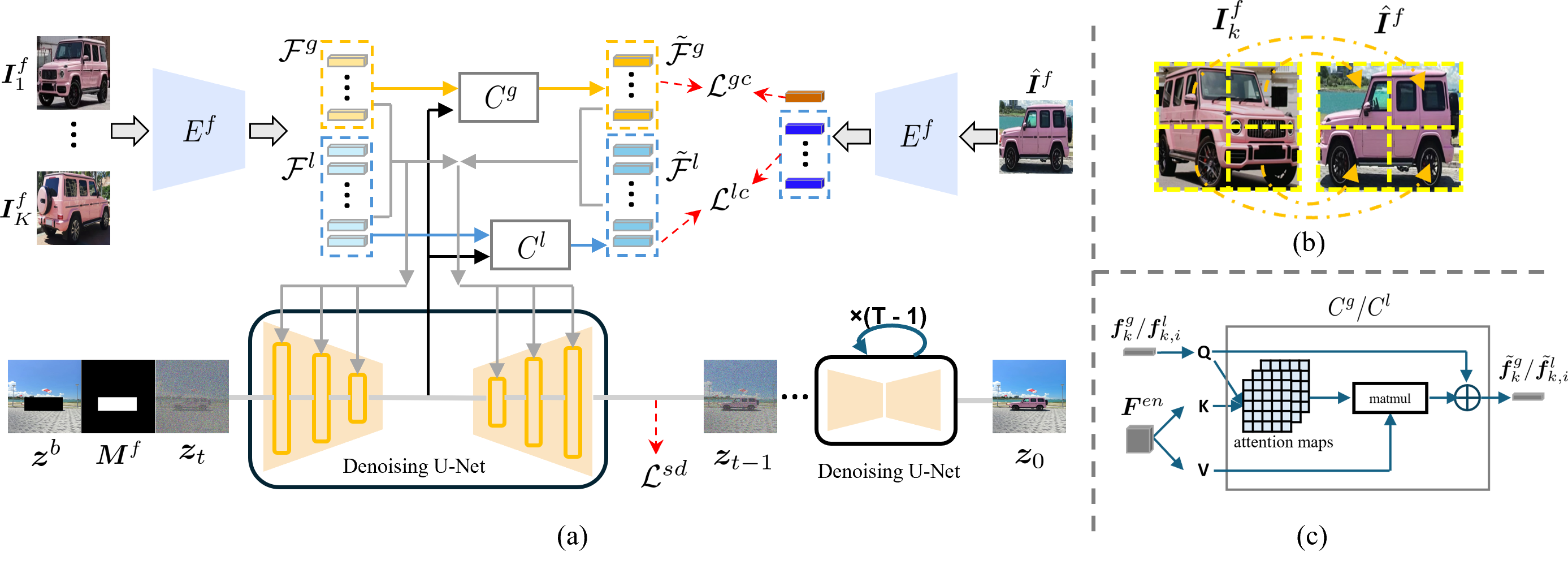} 
\caption{(a) Given multiple foreground reference images, 
we extract their global/local features $\mathcal{F}^g$/$\mathcal{F}^l$, which are passed through the calibration module $C^g$/$C^l$. 
The calibrated features $\tilde{\mathcal{F}}^g$/$\tilde{\mathcal{F}}^l$ are injected into the decoder of denoising UNet. 
(b) Illustration of seeking for the spatial correspondence of local patches between foreground reference $\bm{I}_k^f$ and ground-truth foreground $\hat{\bm{I}}^f$. 
(c) The structure of calibration module $C^g$/$C^l$.
}
\label{fig:pipeline}
\end{figure*}
\subsection{Multi-reference Composition Framework} \label{sec:multi_ref_com}
In this work, we propose a multi-reference generative composition framework, aiming to insert the foreground object at the specified location in the background image based on several reference images of the foreground object. The inserted foreground is expected to maintain the detail information and match the background \emph{w.r.t.} illumination, pose/view, and so on. 
Formally, given a complete background image with a bounding box $\bm{b}$ to place the foreground object and a set of foreground reference images set $\{\bm{I}^f_1, \bm{I}^f_2,..., \bm{I}^f_K\}$, the generated image should resemble the ground-truth image $\hat{\bm{I}}$.

Our model is built upon ObjectStitch~\cite{Song_2023_CVPR} considering its compelling ability to adjust foreground pose/view and generate realistic images. We get the foreground mask $\bm{M}^f$ based on the bounding box $\bm{b}$ and get the masked background $\bm{I}^b$ by erasing the content within $\bm{b}$. 
Following \cite{Song_2023_CVPR}, we first project $\bm{I}^b$ and $\hat{\bm{I}}$  into latent space with VAE encoder, yielding $\bm{z}^b$ and $\bm{z}$ respectively. 
Then, we concatenate $\bm{z}^b$, noisy $\bm{z}_t$, and $\bm{M}^f$ as the input of denoising UNet.
For the $k$-th foreground reference image, we extract its global reference features $\bm{f}_k^g$ and local reference features $\{\bm{f}_{k,i}^l|_{i=1}^N\}$ using $E^f$ which includes the pretrained CLIP encoder~\cite{clip} and an adapter. Global features are the CLS token output with dimension $1\times 1024$ from CLIP encoder, while
local features are the remaining output tokens with dimension  $256\times 1024$ from the last layer.

To obtain the reference features compatible with the background, we design a global calibration module $C^g$ to produce the calibrated global reference features $\tilde{\bm{f}}_k^g$. Similarly, we also design a local calibration module $C^l$ to produce the calibrated local reference features $\{\tilde{\bm{f}}_{k,i}^l|_{i=1}^N\}$. 
We denote the set of global (\emph{resp.}, local) reference features from all reference images as $\mathcal{F}^g$ (\emph{resp.}, $\mathcal{F}^l$).  
Besides, we refer to the calibrated reference features as augmented reference features, and 
denote the set of global (\emph{resp.}, local) augmented reference features from all reference images as $\tilde{\mathcal{F}}^g$ (\emph{resp.}, $\tilde{\mathcal{F}}^l$). 
The original reference features  $\{\mathcal{F}^g, \mathcal{F}^l\}$ are injected into both encoder and decoder of denoising UNet as in \cite{Song_2023_CVPR}. The augmented reference features  $\{\tilde{\mathcal{F}}^g, \tilde{\mathcal{F}}^l\}$ are only injected into the decoder, because the generation of augmented reference features relies on the encoder features of denoising UNet and injecting augmented reference features into the encoder would cause dependency loop.

In the training stage, given a set of training images containing the specific foreground object $\{\bm{I}^o_1, \bm{I}^o_2,..., \bm{I}^o_K\}$,  the foreground images are cropped from training images, followed by geometry and color perturbation. The perturbed foreground images form the set of foreground reference images $\{\bm{I}^f_1, \bm{I}^f_2,..., \bm{I}^f_K\}$. Then, we take one training image $\bm{I}^o_k$ as the ground-truth image $\hat{\bm{I}}$ and prepare the corresponding masked background $\bm{I}^b$, foreground mask $\bm{M}^f$.

The latent $\bm{z}$ of $\hat{\bm{I}}$ is added with $t$-step noise, leading to $\bm{z}_t$. The denoising UNet with parameters $\theta$ is trained using the following objective to predict the added noise:
\begin{equation}
    \label{eqn:loss_ldm}
    \mathcal{L}^{sd} = \mathbb{E}_{\epsilon \sim \mathcal{N}(0,1), t} \| \epsilon - \epsilon_{\theta} \left(\bm{z}^b, \bm{z}_t, \bm{M}^f , \{\mathcal{F}^g, \mathcal{F}^l\}, t  \right) \|^2_2, 
\end{equation}
where $\epsilon$ is the added Gaussian noise and $t$ is the time step ranging from 0 to $T$. 

In the testing stage, given a background image with bounding box, we can obtain $\bm{z}^b, \bm{z}_T, \bm{M}^f$. We can obtain $\mathcal{F}^g, \mathcal{F}^l$ from foreground reference images. Then, we pass through the denoising process to get the denoised latent $\bm{z}_0$, which is mapped back to image space with VAE decoder.

\subsection{Global Reference Feature Calibration} \label{sec:global_cal}

For each global reference feature $\bm{f}^g_k$ (the global feature of foreground reference image $\bm{I}_k^f$), we use global reference feature calibration (GRFC) module to calibrate $\bm{f}^g_k$ to match the ground-truth global feature $\hat{\bm{f}}^g$ (the global feature of ground-truth foreground). Specifically, we crop the foreground $\hat{\bm{I}}^f$ from ground-truth image $\hat{\bm{I}}$ and use foreground encoder to extract its global feature $\hat{\bm{f}}^g$. 

In the GRFC module, we employ the encoder features in denoising UNet to facilitate the calibration process. Since denoising UNet takes the masked background and denoised latent as input, its encoder features should contain the rich information of background and denoised foreground. The background information is helpful to make the calibrated global reference feature compatible with the background. The current denoised foreground may also provide useful hints for the calibration.

Formally, we use each $\bm{f}^g_k$ as query, while the encoder features $\bm{F}^{en}$ in denoising UNet are used as keys and values.
We pass them through a cross-attention layer to produce the calibrated global reference features $\tilde{\bm{f}}^g_k$, which can be formulated as
\begin{equation}
    \label{eqn:calibrate_global}
    \tilde{\bm{f}}^g_k = \textrm{Softmax}\left(\frac{\bm{f}^g_k (\bm{F}^{en}\bm{W}^{gk})^T}{\sqrt{d}}\right)(\bm{F}^{en}\bm{W}^{gv})+\bm{f}^g_k, 
\end{equation}
where $\bm{W}^{gk}, \bm{W}^{gv}$ are projection matrices and $d$ is the dimension of query feature.

The calibrated global reference feature $\tilde{\bm{f}}^g_k$ is forced to match the ground-truth global feature $\hat{\bm{f}}^g$ using the following loss:
\begin{equation}
    \label{eqn:loss_global}
    \mathcal{L}^{gc} = \sum_{k=1}^{K}||\tilde{\bm{f}}^g_k- \hat{\bm{f}}^g||^2.
\end{equation}

The calibrated global reference features $\tilde{\bm{f}}^g_{k}$ form the augmented global reference feature set  $\tilde{\mathcal{F}}^{g}$, which is injected to the decoder of denoising UNet via cross-attention. 

\subsection{Local Reference Feature Calibration} \label{sec:local_cal}
Previous methods~\cite{chen2024anydoor,zhang2023controlcom} have demonstrated that local features of foreground object play an important role in keeping the object details, so we also utilize the local features of foreground reference images. For each local reference feature $\bm{f}^l_{k,i}$ (the $i$-th local feature of foreground reference image $\bm{I}_k^f$), we adopt local reference feature calibration (LRFC) module to calibrate $\bm{f}^l_{k,i}$ to match its corresponding ground-truth local  feature. 

Different from global reference feature calibration, how to obtain the ground-truth local feature is not straightforward. 
Considering that the local feature represents the information of the corresponding local patch, we seek for the spatial correspondence between the patches in foreground reference image and the patches in ground-truth foreground image.  
Given the ground-truth foreground image $\hat{\bm{I}}^f$, we extract its 
local features $\{\hat{\bm{f}}_{k,i}^l|_{i=1}^N\}$. Given the local reference features $\{\bm{f}_{k,i}^l|_{i=1}^N\}$ of the $k$-th reference 
image, for each local reference feature $\bm{f}_{k,i}^l$,   
we find its nearest feature in $\{\hat{\bm{f}}_{k,i}^l|_{i=1}^N\}$ as its ground-truth local feature. 
Specifically, we calculate the similarity between $\{\hat{\bm{f}}_{k,i}^{l}|_{i=1}^{N}\}$ 
and $\{\bm{f}_{k,i}^{l}|_{i=1}^{N}\}$. Based on the $N\times N$ similarity matrix, we can associate the $i$-th patch in the foreground reference image with the most similar $\delta(i)$-th patch in the ground-truth foreground image.

We employ a calibration module $C^l$ to produce the calibrated local reference feature $\tilde{\bm{f}}^l_{k,i}$, which should be close to $\hat{\bm{f}}_{k,\delta(i)}^{l}$. To facilitate the calibration process, similar to Section~\ref{sec:global_cal}, we use the encoder features $\bm{F}^{en}$ in the denoising UNet to provide auxiliary information, because the information of background and denoised foreground could help determine how the local patch should be warped or transformed. 
Formally, we use each local reference feature $\bm{f}_{k,i}^l$ as query, and the encoder features $\bm{F}^{en}$ in the denoising UNet as keys and values. 
We pass them through a cross-attention layer to produce the calibrated local reference feature $\tilde{\bm{f}}^l_{k,i}$, which can be formulated as
\begin{equation}
    \label{eqn:calibrate_local}
    \tilde{\bm{f}}^l_{k,i} = \textrm{Softmax}\left(\frac{\bm{f}^l_{k,i} (\bm{F}^{en}\bm{W}^{lk})^T}{\sqrt{d}}\right)(\bm{F}^{en}\bm{W}^{lv})+\bm{f}^l_{k,i}, 
\end{equation}
where $\bm{W}^{lk}, \bm{W}^{lv}$ are projection matrices and $d$ is the dimension of query feature.

The calibrated local reference features are supervised by 
\begin{equation}
    \label{eqn:loss_local}
    \mathcal{L}^{lc} = \sum_{k=1}^{K}\sum_{i=1}^{N}||\tilde{\bm{f}}_{k,i}^{l}- \hat{\bm{f}}_{k,\delta(i)}^{l}||^2.
\end{equation}

The calibrated local reference features $\tilde{\bm{f}}^l_{k,i}$ form the augmented local reference feature set  $\tilde{\mathcal{F}}^{l}$, which is injected to the decoder of denoising UNet via cross-attention. 

\subsection{Training Strategy} \label{sec:training_strategy}

Our model requires pretraining and few-shot finetuning. 1) We first pretrain our designed GRFC and LRFC using a large-scale training set  (\emph{e.g.}, MVImageNet \cite{yu2023mvimgnet}) which provides multiple images for each object.  The other modules including denoising UNet and VAE are borrowed from the pretrained ObjectStitch \cite{Song_2023_CVPR} model. 2) After pretraining, given a few training images containing a specific foreground object, we finetune the whole model based on these training images. Given test background images with bounding boxes, we can apply the finetuned model to insert the specific foreground object into background. 

\section{Experiments}
\label{sec:experiments}
\subsection{Datasets}
Since multiple reference images are needed, we conduct experiments on two datasets which have multiple reference images for each foreground object.

\noindent\textbf{MureCom}~\cite{lu2023dreamcom}  contains 32 foreground categories. Each foreground category has 3 objects and 20 background images. 
Each object has 5 images with different poses and viewpoints. 
Each background image has a bounding box to specify where the foreground should be inserted.  
For each object, we use its 5 images for few-shot finetuning and 20 background images belonging to its category for evaluation. 

\noindent\textbf{MVImgNet}~\cite{yu2023mvimgnet} contains 222,929 objects from 238 foreground categories. 
Each object has a set of images captured from different camera viewpoints. 
We select one object from each category to form the test objects. 
The images of the remaining objects are used to pre-train GRFC and LRFC as described in Section~\ref{sec:training_strategy}.  
Each test object is associated with 5 images, in which 4 images are 
used for few-shot finetuning and the last image is used for evaluation.

\subsection{Evaluation Metrics} \label{sec:metrics}
We use DINO Score~\cite{caron2021emerging} to assess the fidelity of generated foreground. Since MVImgNet has ground-truth foreground while MureCom does not, DINO score is calculated based on generated foreground and ground-truth foreground (\emph{resp.}, its nearest foreground reference) on MVImgNet (\emph{resp.}, MureCom).  For the background, we use SSIM~\cite{wang2004image} to evaluate background preservation. We choose FOSScore~\cite{zhang2023foreground} to evaluate the pose/view compatibility between foreground and background.
Quality Score (QS)~\cite{wang2004image} is used to evaluate the overall quality of generated images. 

For user study, following previous works~\cite{chen2024anydoor,Song_2023_CVPR}, we invite 50 participants to evaluate the generated images from three aspects:
\emph{fidelity} which measures the foreground detail preservation, \emph{compatibility} which measures the pose/view compatibility between foreground and background,
\emph{quality} which measures the overall quality. We select
100 image sets from each dataset for evaluation. Each image set includes the foreground object, the background image, and 3 generated images from each method. We calculate the average ranking for these three metrics, where 1 indicates the best performance and 5 indicates the worst.

\subsection{Implementation Details}
Our training process consists of two stages: pretraining and fine-tuning. 
In the first stage, the model is pretrained on MVImgNet dataset. The number of training epochs is set to 50, with a batch size 64. 
This stage is conducted on 16 V100 GPUs. In the second stage, the model is finetuned using up to five images containing specific foreground objects.
Fine-tuning stage is conducted on 1 A6000 GPU that takes about ten minutes for 150 epochs.
\begin{figure*}[t]
\centering
\includegraphics[width=\linewidth]{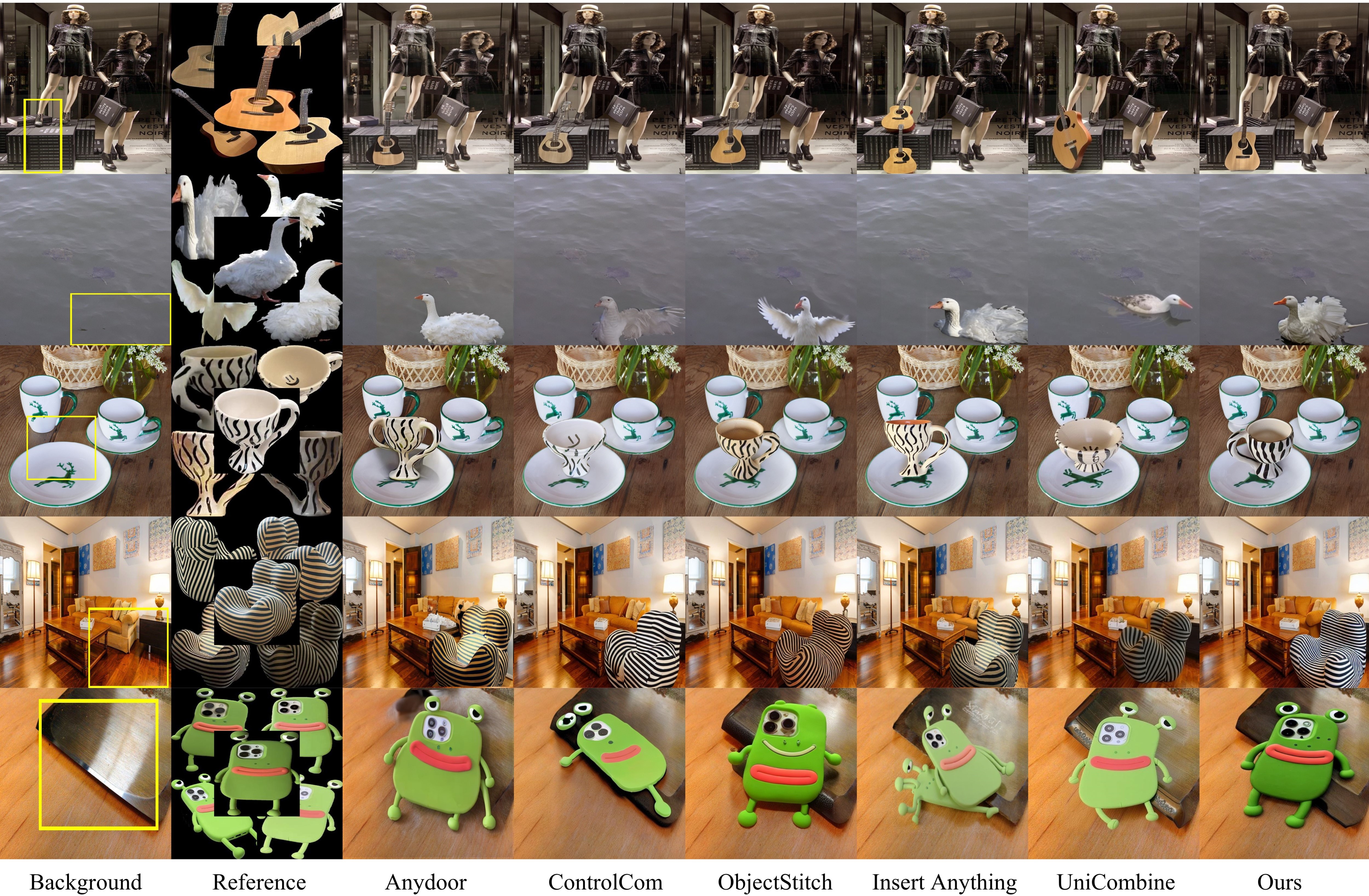} 
\caption{Visual comparison of different methods on MureCom dataset. From left to right, we show background, 5 reference images, the results of  Anydoor~\cite{chen2024anydoor}, ControlCom~\cite{zhang2023controlcom}, ObjectStitch~\cite{Song_2023_CVPR},  Insert Anything~\cite{song2025insert}, UniCombine~\cite{wang2025unicombine} and our CareCom. }
\label{fig:comparison_result}
\end{figure*}
\subsection{Baselines}
We compare our CareCom with recent and open-sourced generative composition methods 
including ObjectStitch~\cite{Song_2023_CVPR}, Anydoor~\cite{chen2024anydoor}, 
ControlCom~\cite{zhang2023controlcom}, Unicombine~\cite{wang2025unicombine} and Insert Anything~\cite{song2025insert}.
Among the baselines, \cite{Song_2023_CVPR,zhang2023controlcom} support multiple reference images.
For these methods, the extension to support multiple reference images is 
similar to our method: the features of multiple reference images are concatenated along the sequence 
dimension and fed into the denoising UNet.
\cite{chen2024anydoor,song2025insert,wang2025unicombine} only supports single reference image. 
For fair comparison, we perform few-shot finetuning for all baselines. 

During inference, for the baselines supporting multiple reference images, we use all reference images.
For the baselines only supporting single reference image, we feed the reference images one by one and get multiple results, from which the best result is selected.

\begin{table}
\centering
\resizebox{\linewidth}{!}{
\begin{tabular}{lccccccc}
\toprule
\multirow{2}*{Method} & \multicolumn{4}{c}{Metrics} &\multicolumn{3}{c}{User Study}\\
\cmidrule(lr){2-5}  \cmidrule(lr){6-8} 
&$\mathrm{DINO_{fg}}$$\uparrow$ &$\mathrm{SSIM_{bg}}$$\uparrow$ & FOSScore$\uparrow$
& QS$\uparrow$  & Fidelity$\downarrow$ & Compatibility$\downarrow$ & Quality$\downarrow$ \\ 
\midrule
ControlCom~\shortcite{zhang2023controlcom}    &  64.92         & 0.858 & 0.856 & 42.27 & 4.41 & 3.54 & 4.10 \\
Anydoor~\shortcite{chen2024anydoor}           & 68.65 & 0.857 & 0.815 & 43.40 & 3.01 & 4.12 & 3.81 \\
ObjectStitch~\shortcite{Song_2023_CVPR}      & 65.04          & 0.854 & 0.867 & 44.39 & 3.48 & 2.72 & 2.95\\
Insert Anything~\shortcite{song2025insert} & \textbf{68.78} & 0.854 & 0.822 & 45.40 & \textbf{2.95} & 4.02 & 3.78 \\ 
UniCombine~\shortcite{wang2025unicombine} & 65.72 & 0.856 & 0.819  & 44.87 &  4.11 & 4.39 & 3.94 \\ 
\midrule
CareCom             & 68.60          & \textbf{0.859} & \textbf{0.883} & \textbf{47.07} & 3.04 & \textbf{2.21} & \textbf{2.42} \\
\bottomrule
\end{tabular}
}
\caption{Quantitative comparison on MureCom dataset. The best results are highlighted in
boldface.}
\label{tab:baseline_MureCom}
\end{table}

\begin{table}
\centering
\resizebox{\linewidth}{!}{
\begin{tabular}{lccccccc}
\toprule
\multirow{2}*{Method} & \multicolumn{4}{c}{Metrics} &\multicolumn{3}{c}{User Study}\\
\cmidrule(lr){2-5}  \cmidrule(lr){6-8} 
&$\mathrm{DINO_{fg}}$$\uparrow$ & $\mathrm{SSIM_{bg}}$$\uparrow$ & FOSScore$\uparrow$ & QS$\uparrow$  & Fidelity$\downarrow$ & Compatibility$\downarrow$ & Quality$\downarrow$ \\ 
\midrule
ControlCom~\shortcite{zhang2023controlcom}   & 63.93 & 0.857 & 0.824 & 37.72 & 4.49 & 3.56 & 4.40 \\
Anydoor~\shortcite{chen2024anydoor}      & 69.56 & 0.853 & 0.791 & 40.85 & 2.90 & 4.45 & 3.72 \\
ObjectStitch~\shortcite{Song_2023_CVPR} & 66.21 & 0.857 & 0.841 & 42.73 & 3.39 & 2.80 & {2.98}\\
Insert Anything~\shortcite{song2025insert} & \textbf{69.88} & 0.857 & 0.804 & 41.22 & \textbf{2.88} & 3.92 & 3.79 \\
UniCombine~\shortcite{wang2025unicombine} &  64.22 & 0.853 & 0.812 & 41.86 & 4.43 & 3.93 & 3.66 \\
\midrule
CareCom         & 69.49 & \textbf{0.858} & \textbf{0.874} & \textbf{45.79} & 2.91 & \textbf{2.34} & \textbf{2.45}\\
\bottomrule
\end{tabular}
}
\caption{Quantitative comparison on MVImgNet dataset. The best results are highlighted in
boldface.}
\label{tab:baseline_MVImgNet}
\end{table}

\subsection{Quantitative Comparison and User Study}

We evaluate different approaches on MureCom and MVImgNet datasets. The results are reported in~\tabref{tab:baseline_MureCom} and~\tabref{tab:baseline_MVImgNet} respectively. The quantitative metrics and user study details have been introduced in Section~\ref{sec:metrics}. 

Anydoor and Insert Anything achieve high DINO scores on both datasets, which shows their ability to preserve foreground details. However, they exhibit obvious copy-and-paste effect and lack the ability to adjust pose/view according to background, indicated by
lower FOSScore. In contrast, although our method performs slightly worse than them on DINO score, it outperforms all the methods for all other metrics.

Based on user study results in \tabref{tab:baseline_MureCom} and~\tabref{tab:baseline_MVImgNet}, consistent with quantitative results, our method slightly underperforms Anydoor and Insert Anything in terms of fidelity, but it significantly outperforms other methods in compatibility and also achieves better overall image quality.

\subsection{Visual Comparison}
In~\figref{fig:comparison_result}, we provide visual 
comparison results of different methods on MureCom dataset. 
The baselines~\cite{zhang2023controlcom,Song_2023_CVPR} and our method use five reference images. 
Since AnyDoor, Insert Anything and UniCombine only support single reference image, we use each of the five reference images separately as input and select the visually best result. 

It can be observed that 
the images generated by~\cite{chen2024anydoor,song2025insert} exhibit noticeable copy-paste artifacts from the reference images, 
resulting in incompatible lighting and perspective between 
foreground and background. 
ControlCom~\cite{zhang2023controlcom} performs well in composite images with 
simple foreground objects but tends to produce some artifacts 
when the foreground objects are complex. ObjectStitch \cite{Song_2023_CVPR} retains most of the foreground information, but still misses or alters some details (\emph{e.g.}, the pattern on the guitar in row 1, unrealistic horse legs in row 2). 
UniCombine~\cite{wang2025unicombine} fails to preserve the detailed information of the foreground 
objects effectively. In contrast, our method is adept at preserving the foreground details and simultaneously adjusting the foreground pose/view to fit the background.

\begin{figure*}[t]
\centering
\includegraphics[width=0.95\linewidth]{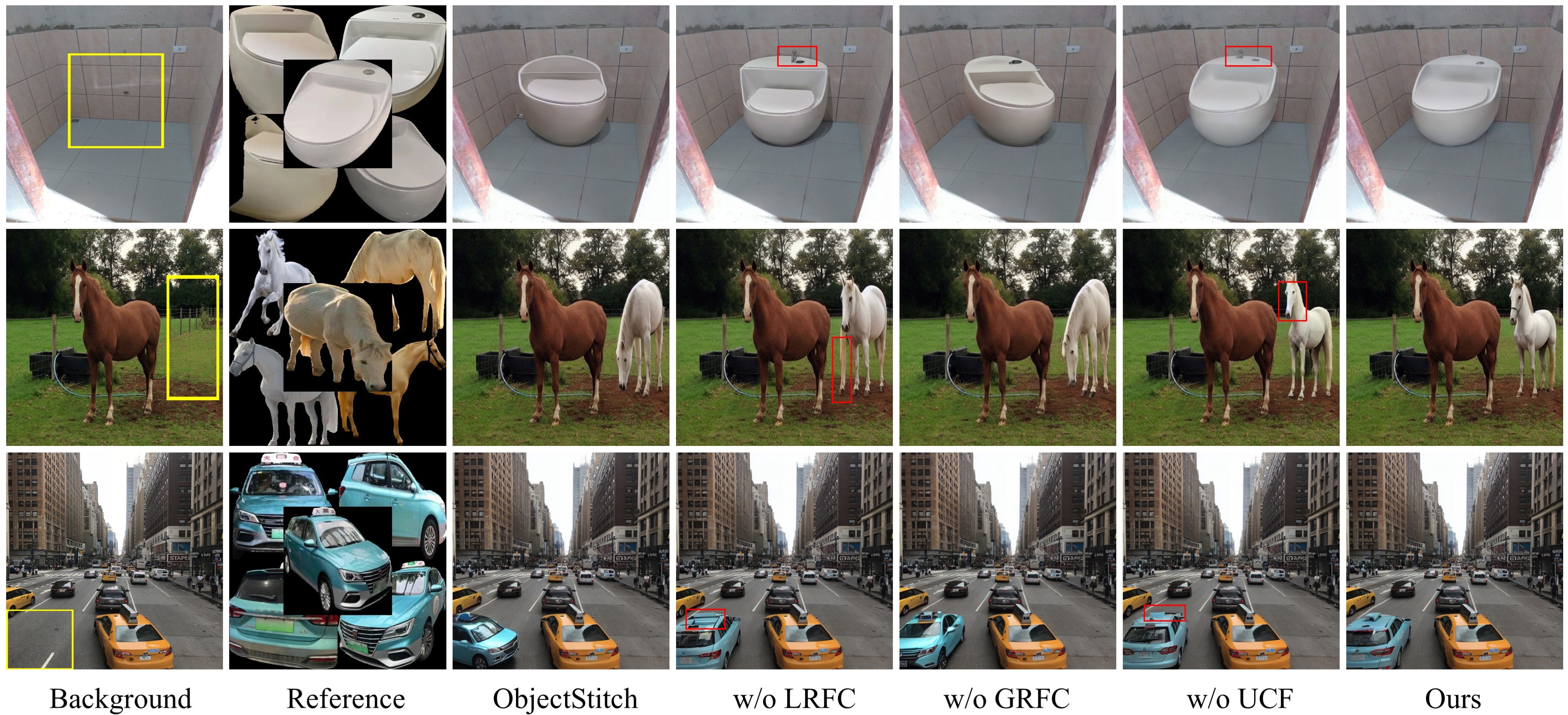} 
\caption{Ablation study of our GRFC/LRFC modules and uncalibrated features (UCF). From left to right, we show background image, five reference images, the results of ObjectStitch, three ablated versions of our method, and our full method. The red boxes indicate the changed details.}
\label{fig:ablation_MureCom}
\end{figure*}

\subsection{Ablation Study}
In this section, we study the impact of global and local feature calibration. 
We progressively add the proposed modules and report the results on MureCom dataset in Table~\ref{tab:ablation}.

The first row shows the results of original ObjectStitch without reference feature calibration. In the second (\emph{resp.}, third) row, we add global (\emph{resp.}, local) reference feature calibration. It can be seen that all four metrics are improved, indicating the enhancement of detail preservation and overall quality.
In the fourth row, we only inject calibrated features into the decoder of denoising UNet. By comparing the first row and the fourth row, we can see that only using calibrated feature can achieve satisfactory performance, which justifies the effectiveness of calibrated features. In the last row, we report the results of our full method. 
Our full method achieves the best results for all metrics. 

\begin{table}[t]
\centering
\resizebox{\linewidth}{!}{
\begin{tabular}{llllllll}
\toprule
GRFC & LRFC & UCF & CF &$\mathrm{DINO_{fg}}$$\uparrow$ 
&$\mathrm{SSIM_{bg}}$ $\uparrow$ & FOSScore$\uparrow$ & QS$\uparrow$ \\ 
\midrule
     &  & \cmark &  & 65.04 & 0.854 & 0.867 & 44.39 \\

\cmark &  & \cmark & \cmark & 65.21 & 0.856 & 0.871 & 45.72 \\  
     & \cmark & \cmark & \cmark & 66.64 & 0.857 & 0.873 & 46.95 \\
\cmark & \cmark &  & \cmark & 66.73 & 0.856 & 0.874 & 46.87 \\
\cmark & \cmark & \cmark & \cmark & \textbf{68.60} & \textbf{0.859} & \textbf{0.883} & \textbf{47.07} \\
\bottomrule
\end{tabular}
}
\caption{Ablation study of the impact of global reference feature 
calibration module (GRFC), local reference feature calibration module (LRFC), 
and whether to use calibrated features (CF) and uncalibrated features (UCF).}
\label{tab:ablation}
\end{table}
We also provide the visualization results of ablation study in~\figref{fig:ablation_MureCom}. 
Without the LRFC module, the ability to preserve foreground details is degraded.
Without the GRFC module, the ability to adjust the foreground pose/view is impaired. 
Without using the uncalibrated features,
some artifacts can be observed on the foreground object,
probably because that the calibration process causes the detail information loss 
and the original reference features can effectively supplement more details.

\begin{figure}[t]
\centering
\includegraphics[width=\linewidth]{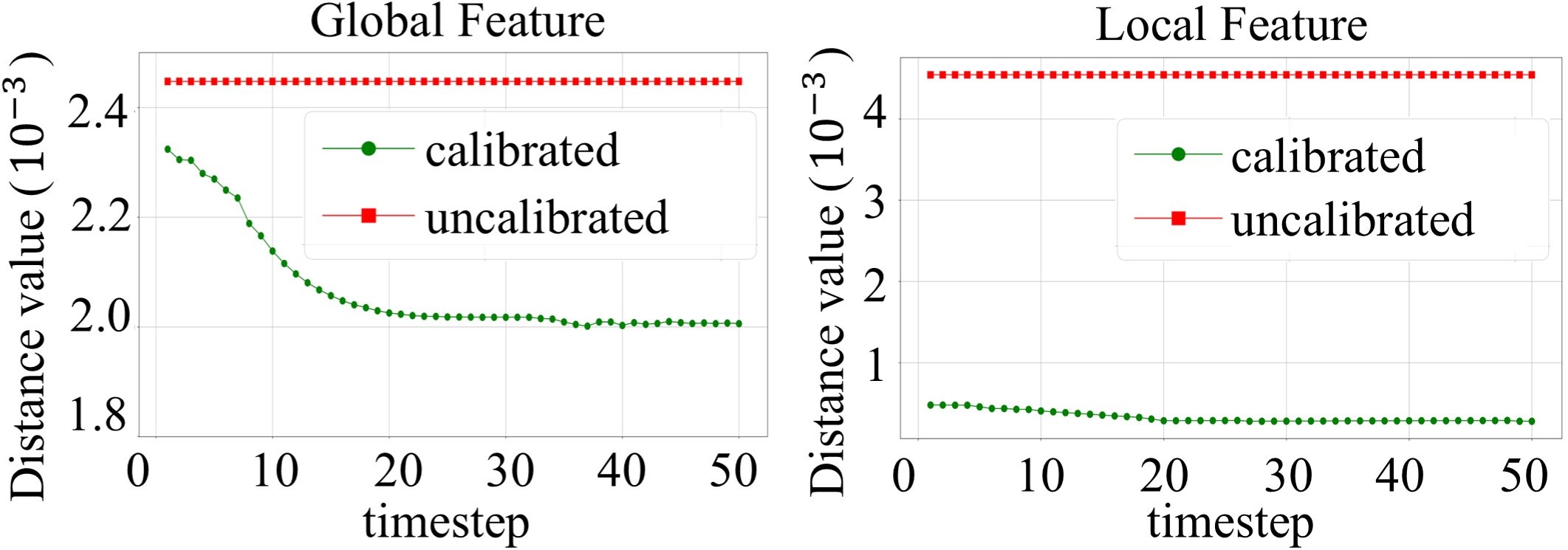} 
\caption{The distance between uncalibrated/calibrated reference features and ground-truth reference features along with the denoising step.}
\label{fig:global_local_loss}
\end{figure}

\begin{figure}[tbp]
\centering
\includegraphics[width=\linewidth]{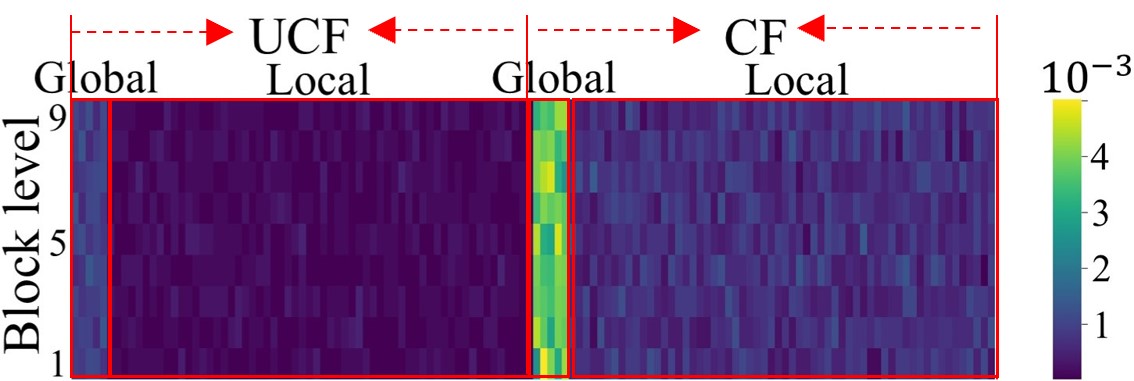} 
\caption{Visualization of cross-attention map in different decoder blocks. Brighter colors indicate larger values.}
\label{fig:attention_map}
\end{figure}

\subsection{Effectiveness of Feature Calibration}
To validate the effectiveness of feature calibration, that is, the calibrated reference features are closer to ground-truth features, 
we calculate the L2 distance between reference features before/after calibration
and ground-truth features on MVImgNet test set, because MVImgNet test set has ground-truth features. 
For each test example, we first extract the global reference features $\{\bm{f}^g_k|_k\}$ and local reference features $\{\bm{f}^l_{k,i}|_{k,i}\}$,
followed by calculating their distances to ground-truth global/local features.
The averaged distances are plotted in red in \figref{fig:global_local_loss}, which serves as the baseline. 
When going through the denoising process, we record the calibrated global reference features $\{\tilde{\bm{f}}^g_k|_k\}$ 
and local reference features $\{\tilde{\bm{f}}^l_{k,i}|_{k,i}\}$ at each timestep, followed by calculating their distances 
to ground-truth global/local reference features.  The averaged distance values are plotted in green in \figref{fig:global_local_loss}.
When compared with uncalibrated reference features, calibrated reference features are progressively getting closer to 
ground-truth reference features as the denoising procedure advances, 
demonstrating the effectiveness of feature calibration.

To further investigate the role of different features in the calibration process, we visualize the cross-attention maps
between decoder features and reference features in different decoder blocks in~\figref{fig:attention_map}. The model
assigns higher attention to the calibrated features (CF), with the calibrated global features receiving the highest attention,
which shows that calibrated features can provide crucial guidance for foreground generation. Uncalibrated features (UCF) are 
assigned relatively low weights, while they still contribute to the generation process to some extent.

\section{Conclusion}

In this paper, we have proposed a multi-reference generative composition framework, which can utilize arbitrary number of
foreground reference images. Under our framework, we have further proposed to calibrate the foreground reference features to be compatible with the background. Comprehensive experiments have verified the effectiveness of our framework equipped with calibrated reference features.  
\section*{Acknowledgements}
The work was supported by the National Natural Science Foundation of China (Grant No. 62471287).

\bibliography{aaai2026}

\end{document}


\maketitle
In this document, we provide additional materials to supplement the main paper. In Section~\ref{sec:num_of_ref}, we will study the impact of the number of reference images. In Section~\ref{sec:finetuning}, we will provide the experiment results on the impact of finetuning. In Section~\ref{sec:efficiency}, we will provide the efficiency comparison between our method and ObjectStitch.
In Section~\ref{sec:visual_murecom}, we will provide more visualization results on MureCom dataset. 
In Section~\ref{sec:visual_mvimg}, we will provide the visualization results on MVImgNet dataset. 
In Section~\ref{sec:single_ref_image}, we will provide visualization results using only one reference image. 
In Section~\ref{sec:ablation_mvimg}, we will report the ablation study results on MVImgNet dataset and show the corresponding visualization results. In Section~\ref{sec:limitation}, we will discuss the limitation of our method. 

\section{Impact of the Number of Reference Images}
\label{sec:num_of_ref}
\begin{figure*}[t]
\centering
\includegraphics[width=\linewidth]{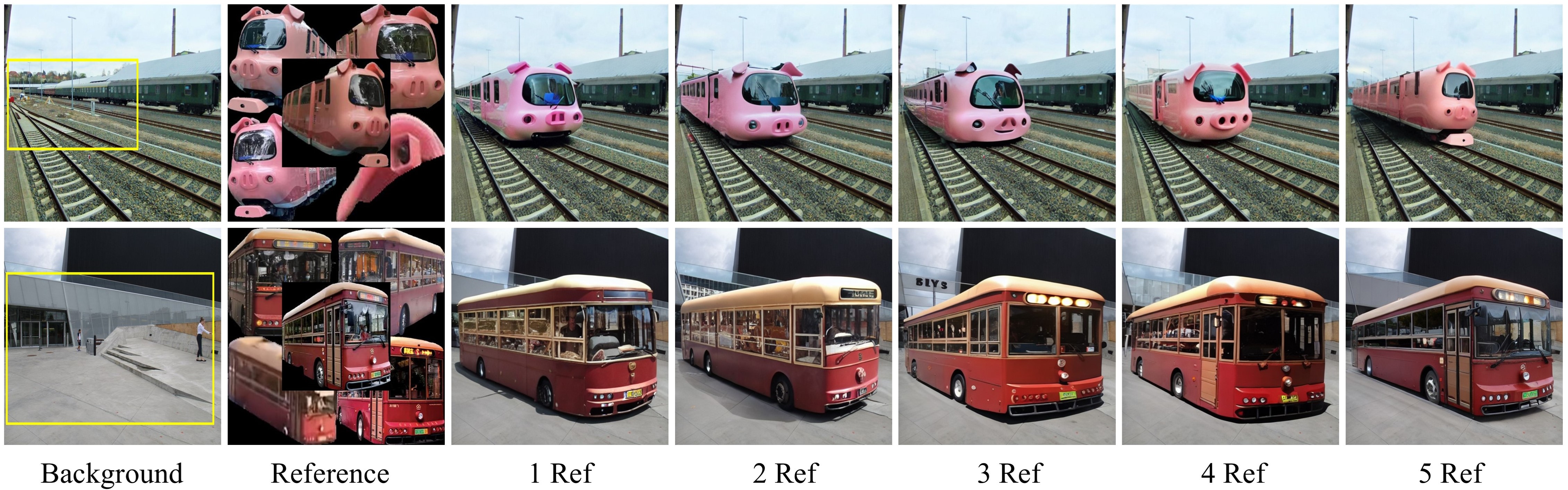} 
\caption{Visual results of our method when using different numbers of reference images on MureCom dataset.}
\label{fig:num_of_fg}
\end{figure*}

In this section, we study the impact of the number of reference images on MureCom dataset. We finetune and test the model using varying numbers of reference images, ranging from 1 to 5.
The visualization results are shown in~\figref{fig:num_of_fg}.

When the foreground objects are complex and have rich details, using a small number of 
reference images can hardly preserve the foreground details. 
However, as the number of reference images increases, the quality of generated images is improved significantly. 

Meanwhile, the example in the second row justifies the necessity of using multiple reference images. The bus has a door only on one side. Given only a single reference image captured from the other side, the model fails to generate the door correctly as shown in the third column. When multiple reference images are provided, our method successfully generates the door as shown in the last column. 

\section{Object-specific Few-shot Finetuning}
\label{sec:finetuning}
Note that for all methods, we perform few-shot finetuning based on a few training images with a specific foreground object. In this section, we show that few-shot finetuning is necessary to achieve both foreground fidelity and foreground-background compatibility. 
We compare our method with ObjectStitch and Anydoor on MureCom dataset, with the results
reported in~\tabref{tab:obj_finetuning}.  ObjectStitch is representative for those methods which are strong in pose/view adjustment but weak in detail preservation. Anydoor is representative for those methods which are strong in detail preservation but weak in pose/view adjustment. 

Without few-shot finetuning, Anydoor achieves the highest foreground fidelity with the DINO score of 67.36, while ObjectStitch and
our method are relatively low. 
The pose/view compatibility between foreground and background can be reflected through the FOSScore.
Due to the copy-and-paste issue, the FOSScore of Anydoor is much lower, 
while ObjectStitch and our method perform better. 

After finetuning, most results are improved to some extent. 
However, Anydoor still exhibits poor pose/view compatibility between 
foreground and background, because its way to inject reference information restricts the flexibility to adjust pose/view.  
In contrast, our method leverages 
calibration modules to correct the pose of reference object, significantly improving compatibility and achieving a FOSScore of 0.883.
And the increase of DINO score demonstrates the foreground fidelity of images generated by ObjectStitch and our method
greatly benefits from few-shot finetuning.


\begin{table}
\centering
\resizebox{\linewidth}{!}{
\begin{tabular}{lcccccccc}
\toprule
\multirow{2}*{Method} &  &\multicolumn{4}{c}{Metrics}\\
\cmidrule(lr){3-6}  
& &$\mathrm{DINO_{fg}}$$\uparrow$ &$\mathrm{SSIM_{bg}}$$\uparrow$ & FOSScore$\uparrow$ & QS$\uparrow$  \\ 
\midrule
ObjectStitch~\cite{Song_2023_CVPR} &        &61.66 & 0.851 & 0.858 & 40.26  \\  
Anydoor~\cite{chen2024anydoor}      & w/o finetune & \textbf{67.36} & \textbf{0.856} & 0.814 & 42.45  \\
Ours         &        & 63.49 & 0.855 & \textbf{0.872} & \textbf{42.92}  \\
\midrule
ObjectStitch~\cite{Song_2023_CVPR} &   & 65.04 & 0.854 & 0.867 & 44.39  \\
Anydoor~\cite{chen2024anydoor}     &finetune & \textbf{68.65} & 0.857 & 0.815 & 43.40 \\
Ours         &   & 68.60 & \textbf{0.859} & \textbf{0.883} & \textbf{47.07} \\
\bottomrule
\end{tabular}
}
\caption{The results of different methods without using or using few-shot finetuning.}
\label{tab:obj_finetuning}
\end{table}
\begin{table}[t]
    \centering
    \resizebox{\linewidth}{!}{
      \begin{tabular}{lcccc}
        \toprule
        Methods &GPU memory(MB) & Parameters(B)  & inference time(s) \\
        \midrule
        Anydoor~\cite{chen2024anydoor} & 18208 & 2.5 & 9 \\
        ControlCom~\cite{zhang2023controlcom} & 13622 & 1.6 & 6.5 \\
        ObjectStitch~\cite{Song_2023_CVPR} & 11352 & 1.31  & 2.27  \\
        UniCombine~\cite{wang2025unicombine} & 35668 & 11.92 & 9 \\
        Insert Anything~\cite{song2025insert} & 37028 & 18.11 & 37 \\
        Ours         & 11470 & 1.33  & 2.61 \\
        \bottomrule
      \end{tabular}
      }
      \caption{Efficiency comparison of different baselines and our method.}
    \label{tab:Efficiency}
\end{table}

\section{Efficiency Comparison}
\label{sec:efficiency}
We compare the computational efficiency of our model with all the baselines. 
In~\tabref{tab:Efficiency}, we report the GPU 
memory usage, parameter count, and inference time. It can be seen that the GPU memory cost and parameter count of
our method are marginally increased compared to ObjectStitch.  For inference time, we test 100 images on a single A6000 GPU and calculate the average. 
Our method has comparable inference time with ObjectStitch. 
Compared with other baselines, our approach significantly outperforms them in all three aspects.
\begin{figure*}[ht]
\centering
\includegraphics[width=\linewidth]{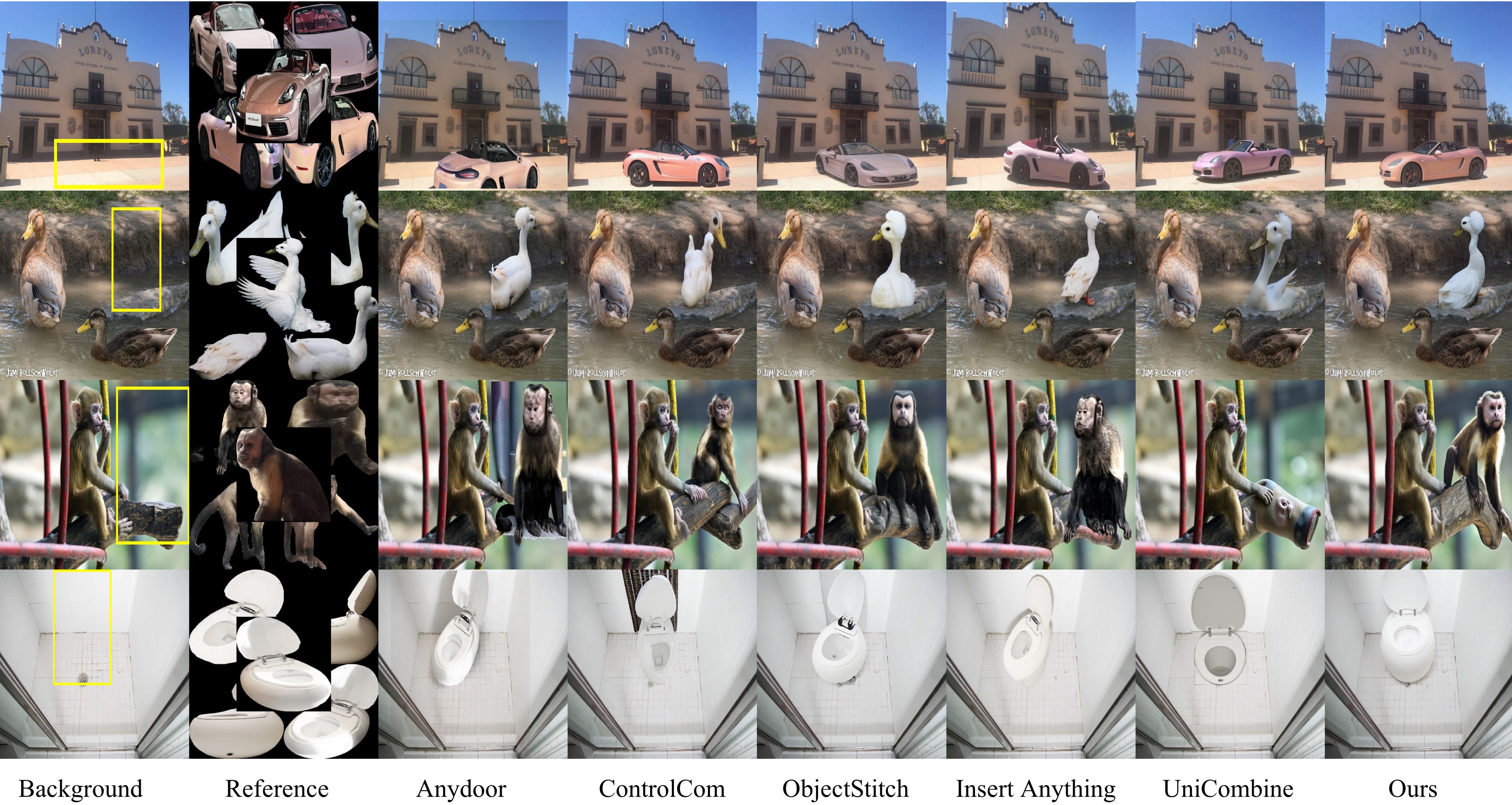} 
\caption{Visual comparison of different methods on Murecom dataset. From left to right, we show background, 5 reference images, the results of Anydoor~\cite{chen2024anydoor}, ControlCom~\cite{zhang2023controlcom}, ObjectStitch~\cite{Song_2023_CVPR},  Insert Anything~\cite{song2025insert}, UniCombine~\cite{wang2025unicombine} and our method. }
\label{fig:vis_murecom_supp} 
\end{figure*}

\section{More Visualization Results on MureCom} \label{sec:visual_murecom}
More visualization results on MureCom dataset are shown in~\figref{fig:vis_murecom_supp}. We can see that our method can adjust the foreground pose and view while preserving the foreground details, notably outperforming the other methods. {The results of Anydoor and Insert Anything look like pasting the reference image on the background, so the adjustment of foreground pose/view is limited and thus the foreground does not appear naturally in the background. Although ControlCom attempts to adjust the foreground pose/view, the generated foregrounds may be deformed and have low quality (\emph{e.g.}, row 2, 4). UniCombine is weak in preserving the foreground details and it sometimes fails to place the foreground image at the specific position.} ObjectStitch is a competitive baseline, but our method can generate more harmonious and realistic results (\emph{e.g.}, row 1).  Moreover, the foreground could interact with the background more naturally and vividly (\emph{e.g.}, row 3). 

\section{Visualization Results on MVImgNet} \label{sec:visual_mvimg}
In this section, we provide visual comparison results of different methods on MVImgNet dataset. 

\begin{figure*}[ht]
\centering
\includegraphics[width=\linewidth]{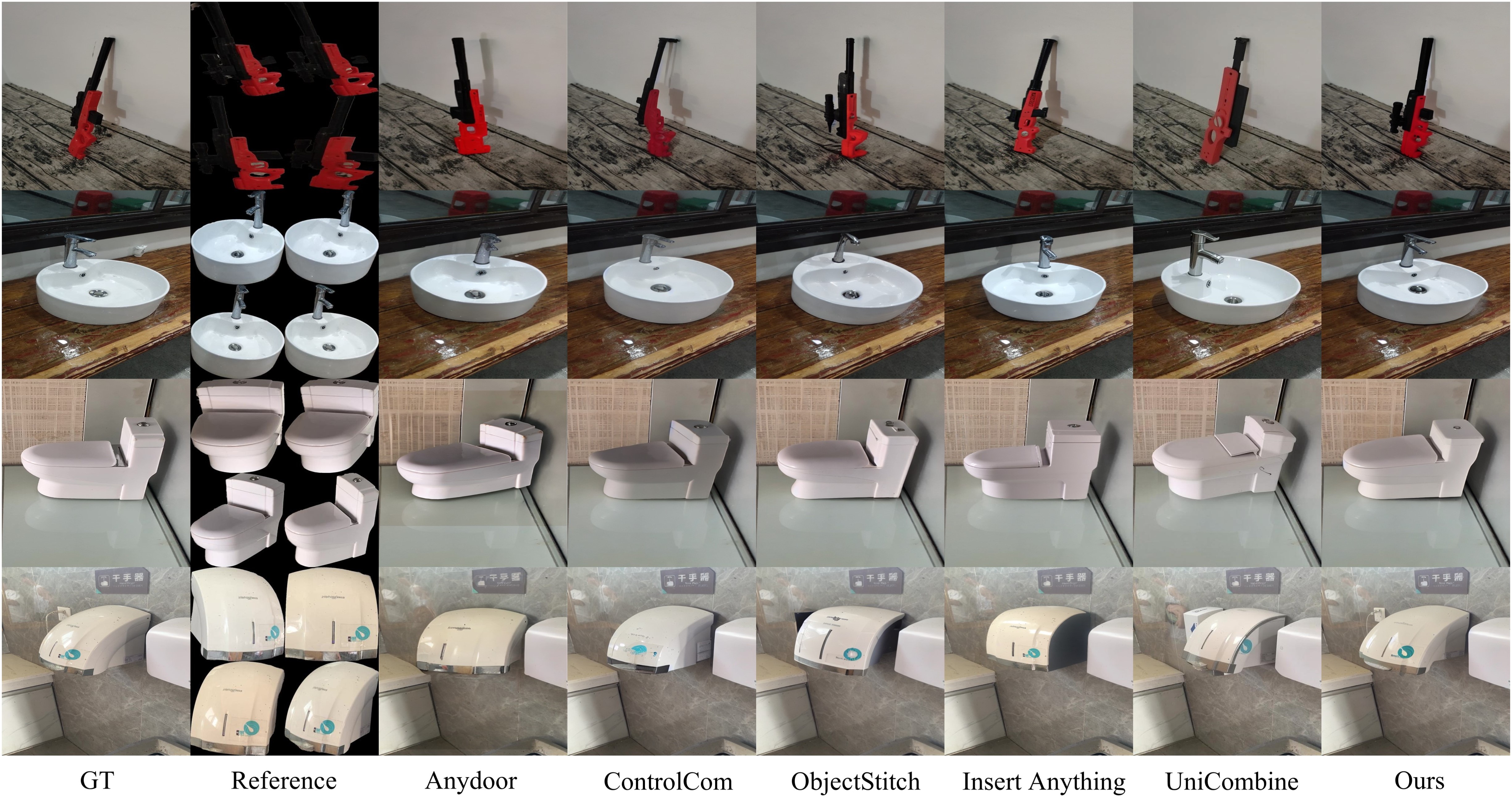} 
\caption{Visual comparison of different methods on MVImgNet dataset. From left to right, we show GT images, 4 reference images, the results of Anydoor~\cite{chen2024anydoor}, ControlCom~\cite{zhang2023controlcom}, ObjectStitch~\cite{Song_2023_CVPR},  Insert Anything~\cite{song2025insert}, UniCombine~\cite{wang2025unicombine} and our method. }
\label{fig:vis_mvimgnet}
\end{figure*}

As shown in~\figref{fig:vis_mvimgnet}, the images generated by Anydoor and Insert Anything exhibit noticeable copy-and-paste artifacts. 
It can only place the foreground object onto the background with the same pose, failing to adjust the foreground object based on the background image and the viewpoint. 
ControlCom and ObjectStitch can leverage information from multiple reference images to adjust the pose of the foreground object to some degree. 
However, they suffer from significant detail loss and still exhibit inconsistencies between the foreground object and background. {UniComine 
can only utilize one reference image and the it can not preserve the details of 
the foreground image.}
In contrast, our method performs well in these two aspects.
It can adjust the pose of foreground object 
according to the background 
while preserving the details of the object.

\section{Visualization Results Using one Reference Image}
\label{sec:single_ref_image}
\begin{figure*}[ht]
\centering
\includegraphics[width=\linewidth]{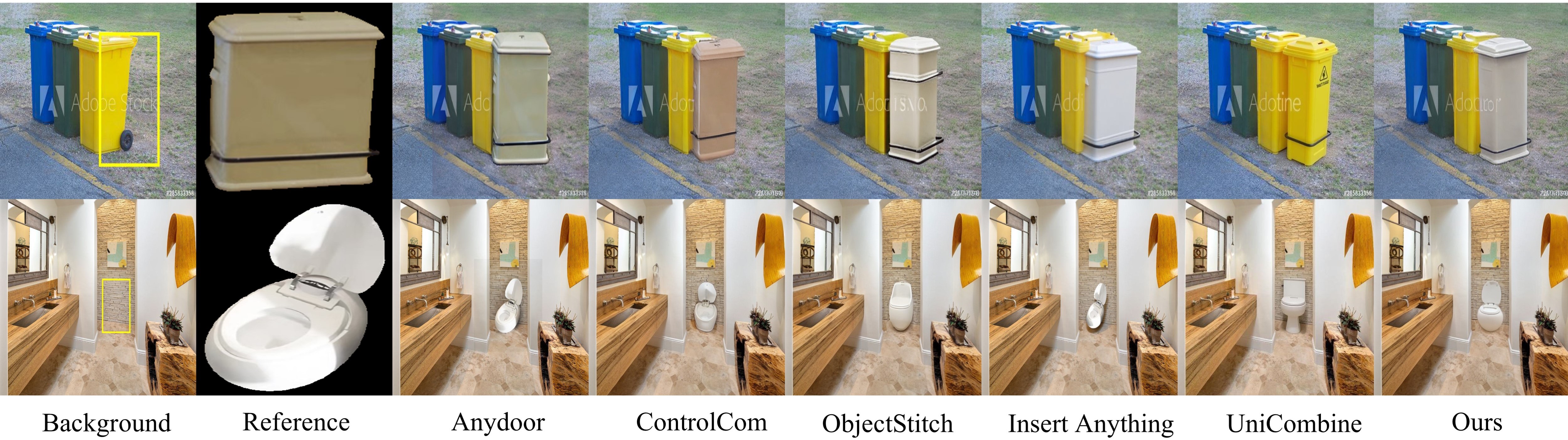} 
\caption{Visual comparison of different methods using single reference image. From left to right, we show background, one reference image, the results of Anydoor~\cite{chen2024anydoor}, ControlCom~\cite{zhang2023controlcom}, ObjectStitch~\cite{Song_2023_CVPR},  Insert Anything~\cite{song2025insert}, UniCombine~\cite{wang2025unicombine} and our method.}
\label{fig:single_ref}
\end{figure*}

{We also compare the results of using only a single reference image.
As shown in~\figref{fig:single_ref}, AnyDoor and Insert Anything fail to adjust the pose of the 
foreground object, while ObjectStitch, ControlCom, and UniCombine struggle to preserve
foreground object details. Our method can do well in both aspects.
Although our method is designed for multiple reference images, it still
outperforms all other methods even when only a single reference image is provided.}
\section{Ablation Study on MVImgNet} \label{sec:ablation_mvimg}

\begin{table}
\centering
\resizebox{\linewidth}{!}{
\begin{tabular}{llllllll}
\toprule
GRFC & LRFC & UCF & CF &$\mathrm{DINO_{fg}}$$\uparrow$ 
& $\mathrm{SSIM_{bg}}$$\uparrow$ & FOSScore$\uparrow$ & QS$\uparrow$ \\ 
\midrule
 &   & \cmark &   & 66.21 & 0.857 & 0.841 & 42.73 \\

\cmark &   & \cmark & \cmark & 66.53 & 0.857 & 0.862 & 43.78 \\  
 & \cmark & \cmark & \cmark & 68.14 & 0.857 & 0.854 & 44.16 \\
\cmark & \cmark &   & \cmark & 67.62 & 0.856 & 0.864 & 44.82 \\
\cmark & \cmark & \cmark & \cmark & \textbf{69.49} & \textbf{0.858} & \textbf{0.874} & \textbf{45.79} \\

\bottomrule
\end{tabular}
}
\caption{Ablation study of the impact of global reference feature 
calibration module (GRFC), local reference feature calibration module (LRFC), 
and whether to use calibrated features (CF) and uncalibrated features (UCF).}
\label{tab:ablation_mvimgnet}
\end{table}

\begin{figure*}[ht]
\centering
\includegraphics[width=\linewidth]{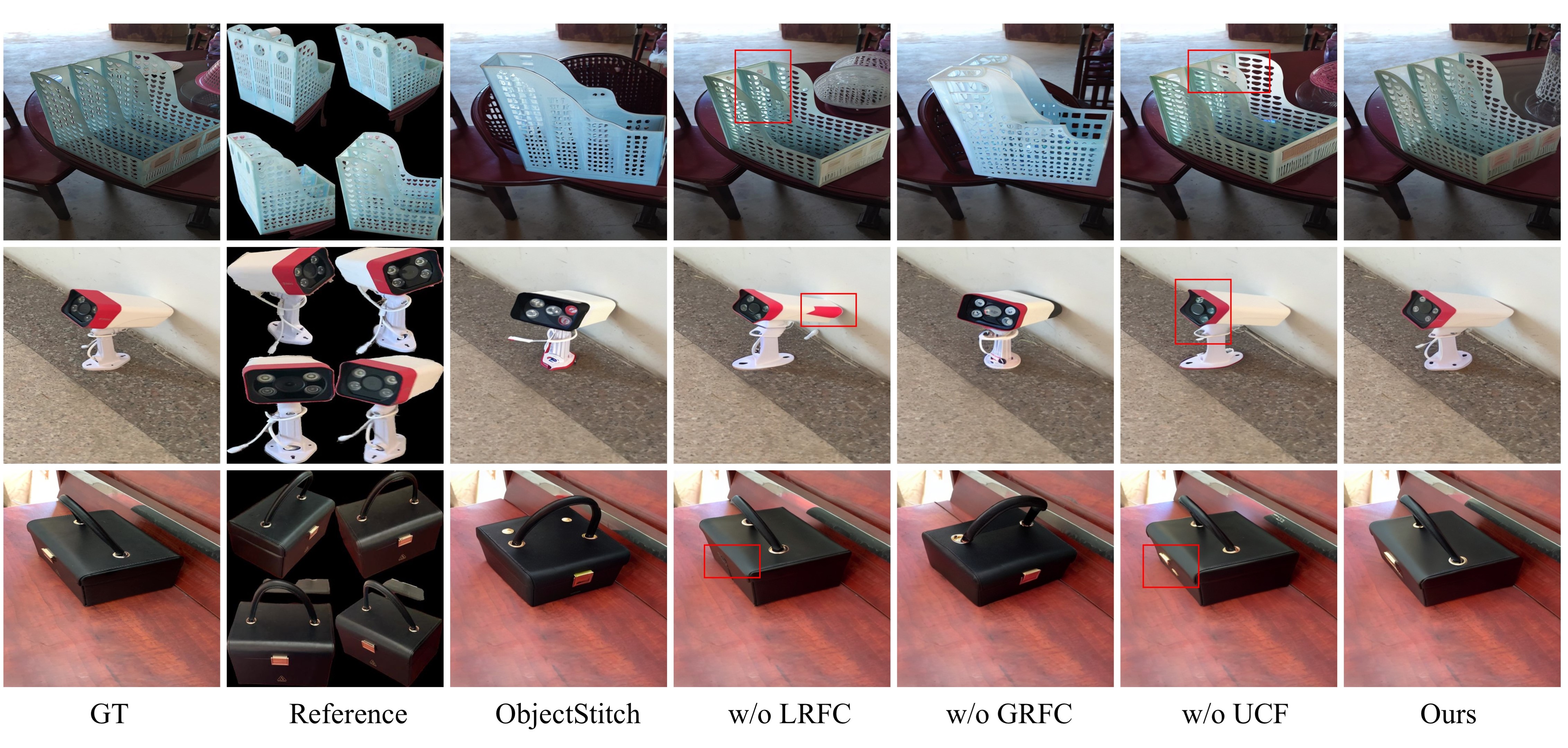} 
\caption{Ablation study of our GRFC and LRFC module. From left to right, we show GT images, five reference images, the results of ObjectStitch, three ablated versions of our method, and our full method. The red boxes indicate the changed details.}
\label{fig:ablation_mvimgnet}
\end{figure*}


In the main paper, we conduct ablation studies on MureCom dataset. 
Here, we conduct the ablation studies on MVImgNet dataset. As in the main paper, We demonstrate the effectiveness of each module in our method.

The quantitative results are shown in~\tabref{tab:ablation_mvimgnet}. 
We observe that after removing the LRFC module, 
the DINO score decreases, indicating a certain loss of detail in the foreground object. 
The FOSScore also shows slight decline. 
When the GRFC module is removed, the DINO score of the object does not decrease significantly.
However, the FOSScore suffers from more substantial drop. 
When our model excludes the uncalibrated features,
all metrics show a certain degree of decline, yet the performance still surpasses those of ObjectStitch, which
verifies the critical role of the calibrated features.

Visualization results of the ablation study are shown in~\figref{fig:ablation_mvimgnet}. 
It can be seen that after removing the GRFC module, 
the details of the foreground object can be well preserved,
but the pose of generated foreground may not align well with the background (\emph{e.g.}, row 3).  
When removing the LRFC module, the foreground object can adjust its view/pose according to the background, but there is a certain degree of loss in detail information.
When we only use calibrated features, there are some artifacts on the foreground object. 
Notably, the results of our method are the closest to the ground-truth image, which justifies the necessity of each module. In summary, it is necessary to calibrate both global reference features and local reference features. Besides, it is beneficial to jointly use uncalibrated features and calibrated features.

\begin{figure*}[ht]
\centering
\includegraphics[width=\linewidth]{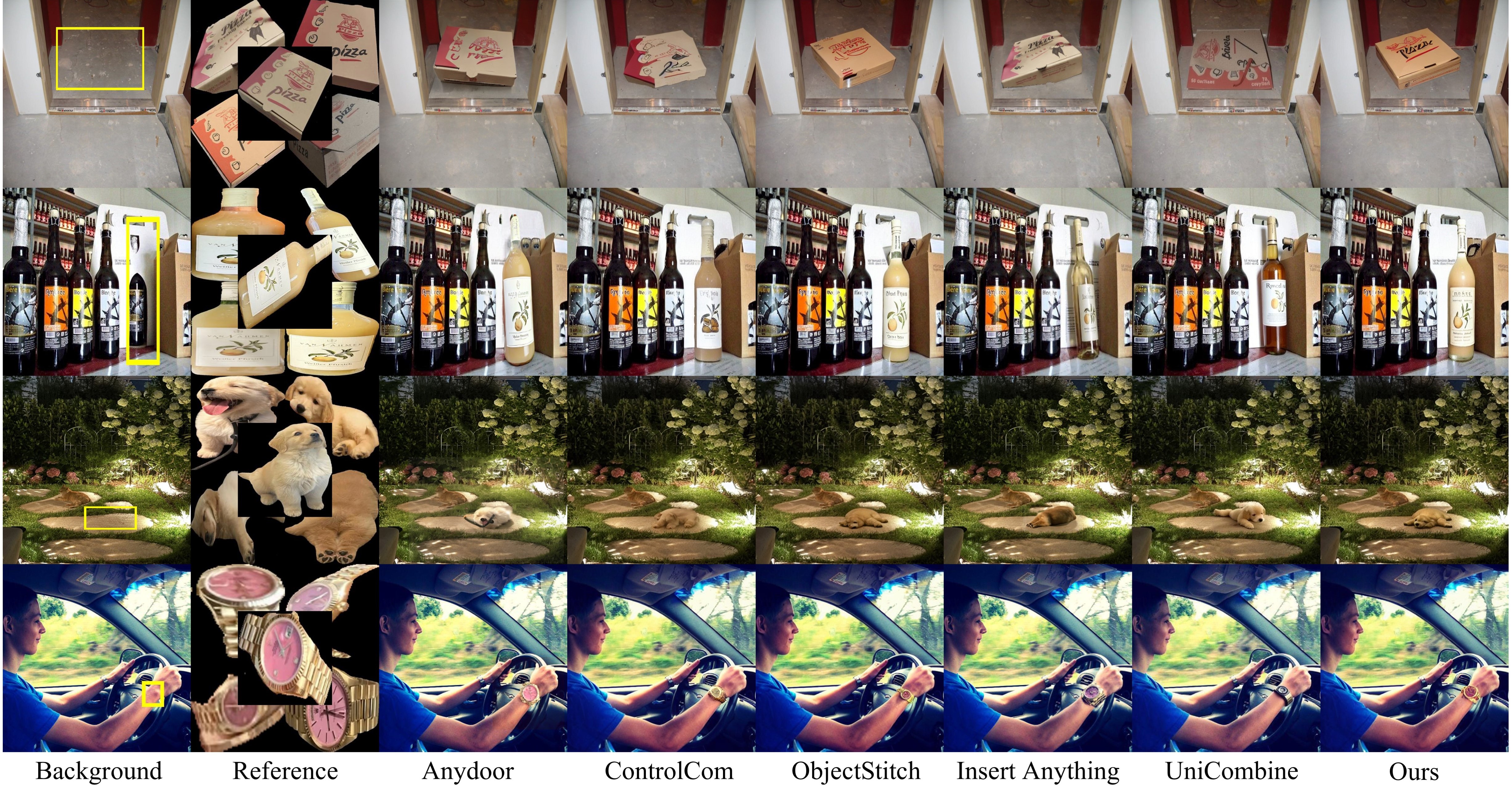} 
\caption{Example failure cases of our method. From left to right, we show background image, 5 reference images, the results of Anydoor~\cite{chen2024anydoor}, ControlCom~\cite{zhang2023controlcom}, ObjectStitch~\cite{Song_2023_CVPR},  Insert Anything~\cite{song2025insert}, UniCombine~\cite{wang2025unicombine} and our method.}
\label{fig:failure_case}
\end{figure*}

\section{Limitation} \label{sec:limitation}
Our proposed method is generally capable of generating images with high fidelity and good compatibility. However, there still exist some issues.

As shown in the first two rows in~\figref{fig:failure_case}.
when the bounding box to place the foreground object is very small,
the details of foreground object in the generated 
image suffer from significant loss. We conjecture that 
the reference features does not cover the information of different scales, so it is difficult to
generate the foreground object at a small scale. One possible solution is taking full advantage of multi-scale reference features. 

As shown the last two rows  of~\figref{fig:failure_case}, another issue is that when the foreground object contains text information, the generated image fails to preserve the text information on the foreground object faithfully. 

Note that the above issues are very challenging for all existing generative composition methods. We also show the results of baselines \cite{chen2024anydoor,zhang2023controlcom,Song_2023_CVPR,song2025insert,wang2025unicombine} in these challenging cases, and they cannot achieve satisfactory performance. 
\clearpage
\newpage
\bibliography{aaai2026}